\documentclass[sigplan]{acmart}

\AtBeginDocument{%
  }

\setcopyright{none}
\usepackage{multirow}
\usepackage{booktabs}
\usepackage{tabularx}
\usepackage[ruled,vlined,linesnumbered]{algorithm2e}

\newcommand{\shaohua}[1]{\textcolor{blue}{@li:#1}}

\begin{document}

\title{\textsc{Trident}: Unifying Guarded Dispatch and Host Execution for PyTorch Triton Workloads}

\author{Jinjie Liu}
\authornote{This work was completed during an internship at the Beijing Academy of Artificial Intelligence.}
\affiliation{\institution{The Chinese University of Hong Kong}\city{Hong Kong}\country{China}}
\email{jjliu26@cse.cuhk.edu.hk}

\author{Xiaoyan Liu}
\authornote{Corresponding author.}
\affiliation{\institution{Beijing Academy of Artificial Intelligence}\city{Beijing}\country{China}}
\email{xyliu01@baai.ac.cn}

\author{Shuhan Zhang}
\affiliation{\institution{Beijing Academy of Artificial Intelligence}\city{Beijing}\country{China}}
\email{liscopye@163.com}

\author{Wenjia Sun}
\affiliation{\institution{Beijing Academy of Artificial Intelligence}\city{Beijing}\country{China}}
\email{sunwenjia04@163.com}

\author{Ruilin Yang}
\affiliation{\institution{Beijing Academy of Artificial Intelligence}\city{Beijing}\country{China}}
\email{rlyang@baai.ac.cn}

\author{Chunlei Men}
\affiliation{\institution{Beijing Academy of Artificial Intelligence}\city{Beijing}\country{China}}
\email{clmen@baai.ac.cn}

\author{Yonghua Lin}
\affiliation{\institution{Beijing Academy of Artificial Intelligence}\city{Beijing}\country{China}}
\email{yhlin@baai.ac.cn}

\author{Shaohua Li}
\affiliation{\institution{The Chinese University of Hong Kong}\city{Hong Kong}\country{China}}
\email{shaohuali@cuhk.edu.hk}

\renewcommand{\shortauthors}{Liu et al.}

\begin{abstract}
User-written Triton kernels enable high-performance GPU computation within PyTorch, but their end-to-end latency can remain dominated by host-side orchestration, especially when device execution is short. Although \texttt{torch.compile} can generate native host wrappers for captured graphs, each invocation still passes through runtime-managed specialization lookup, guard evaluation, and preparation before reaching the wrapper. 

We present \textsc{Trident}, a compiler backend that removes this recurring overhead from the specialization cache-hit path. \textsc{Trident} introduces the \emph{Specialization Cache Module} (SCM), which compiles guarded specialization selection, argument and execution-environment preparation, and host execution for multiple specializations into a single executable module. An invocation enters the SCM once, remains in compiled code when a specialization matches, and returns to Python only when a new specialization must be compiled. Built on Torch-MLIR, \textsc{Trident} lowers guards and host-side orchestration to native code while retaining calls to optimized runtime implementations of supported ATen operators. Our evaluation on two LLMs shows that \textsc{Trident} achieves up to a 1.47$\times$ speedup in model-level end-to-end latency over eager execution and up to 1.68$\times$ over \texttt{torch.compile}.
\end{abstract}

\maketitle

\section{Introduction}
\label{sec:introduction}

With the widespread adoption of large language models~\cite{touvron2023llama,achiam2023gpt,liu2024deepseek,yang2025qwen3}, machine learning frameworks~\cite{pytorch1,abadi2016tensorflow,bradbury2018jax,jia2014caffe,team2016theano} have become an indispensable part of modern software infrastructure. Among these frameworks, PyTorch~\cite{pytorch1} has emerged as a leading platform due to its ease of use, flexible programming model, and strong community support. Building on this eager programming model, PyTorch~2 further introduces TorchDynamo and TorchInductor to enable graph compilation without sacrificing Python flexibility~\cite{pytorch2}. As a representative eager-mode system, PyTorch has become foundational infrastructure for both model research~\cite{wolf2020transformers} and development and production serving stacks~\cite{kwon2023efficient,zheng2024sglang}.


Within the PyTorch ecosystem, Triton has emerged as a widely used backend for customized GPU kernels, enabling users to express high-performance kernels that can be called directly from PyTorch. However, optimized device code alone does not guarantee low end-to-end latency. Each kernel invocation still incurs host-side orchestration overhead, such as buffer allocation, kernel-variant selection, and kernel launch. This mismatch is particularly significant for short kernels, which are common and repeatedly executed in large-model workloads~\cite{kwon2023efficient,zheng2024sglang}. As GPU throughput improves, such kernels become shorter, and host-side kernel launch and orchestration overhead can become comparable to or even exceed device execution~\cite{ghosh2026grace}. Consequently, reducing host-side overhead is essential to fully realize the performance of customized Triton kernels~\cite{pytorch2}.

One approach to reducing these host-side costs is to manually adopt CUDA Graphs, which capture and replay a fixed sequence of GPU operations. Their use, however, requires the captured execution to satisfy constraints on control flow, memory addresses, and kernel behavior. Meeting these constraints often entails substantial modifications to both host-side calling logic and the kernels themselves~\cite{kwon2023efficient}. Even when capture is feasible, parameter-copying and replay-related overheads can outweigh its benefits and degrade performance~\cite{ghosh2026grace}. These limitations have motivated framework support that applies host-side optimizations transparently while preserving the eager programming interface.

\begin{figure*}[t]
  \centering
  \includegraphics[width=2\columnwidth]{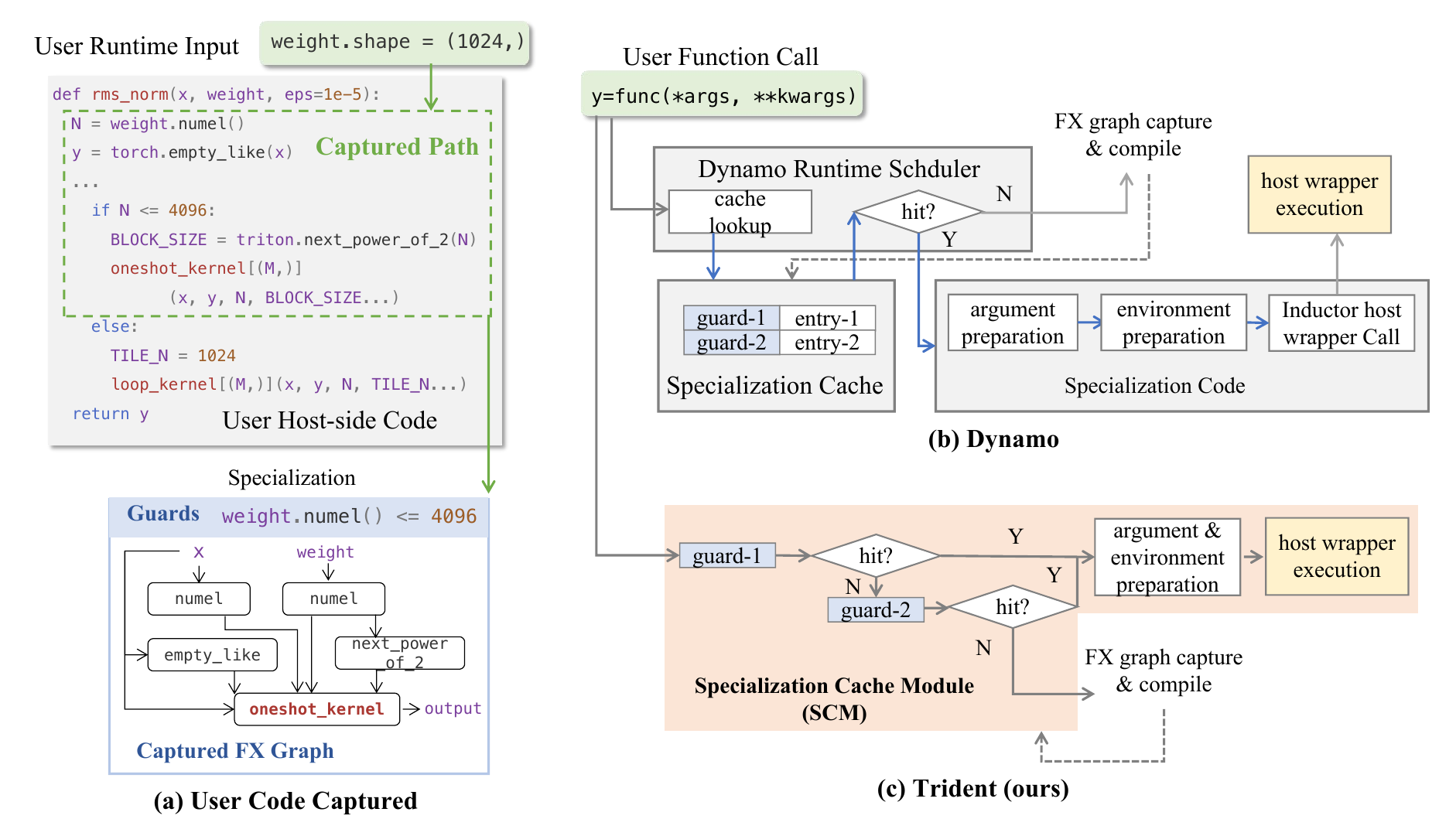}
  \caption{(a)~An \texttt{rms\_norm} example (omitted for clarity) with two Triton JIT kernels, \texttt{oneshot\_kernel} and \texttt{loop\_kernel}. Different input shapes take different host-side paths; Dynamo captures the executed path as a guarded FX-graph specialization.
  (b)~Dynamo runtime-managed execution model, where blue edges mark cross-layer and glue overhead between warm-path steps.
  (c)~\textsc{Trident} SCM-based execution model.}
  \label{fig:intro-guard}
\end{figure*}

PyTorch provides this support using Dynamo for graph capture and Inductor as its default compiler backend~\cite{pytorch2}. As shown in Figure~\ref{fig:intro-guard}(a), Dynamo executes and traces one path through the user's host-side Python code, capturing the tensor operations and Triton kernel calls along that path in an FX graph~\cite{reed2022torch}. During graph capture, Dynamo records guards that describe the input properties and runtime assumptions under which the captured path remains valid. The resulting FX graph and its guards form a \emph{specialization}, which Dynamo caches for subsequent invocations.
Once the operations of a specialization are known, Inductor generates a host wrapper that performs tasks such as memory management, launch-argument preparation, and kernel launch. By default, Inductor emits this wrapper as Python code. Although graph capture removes repeated interpretation of the original user code, executing the generated wrapper still incurs Python dispatch and object-handling overhead. To further reduce this overhead, Inductor can instead generate a C++ wrapper, allowing the host operations within the captured graph to execute as compiled native code.

However, the C++ wrapper covers only the host operations derived from one FX graph, while each invocation still incurs host-side work outside the wrapper.
As illustrated by Figure~\ref{fig:intro-guard}(b), every user function call enters the Dynamo runtime scheduler, which performs a lookup in the specialization cache and evaluates guards to select matching specialization code. The selected specialization code performs argument preparation and environment preparation before calling the Inductor host wrapper. This preparation is necessary to preserve the execution semantics of the Python function when invoking the selected FX graph, but introduces host-side overhead through additional runtime calls and glue code.
This host-side cost can slow down the operator's end-to-end execution, especially for short Triton kernels.


These costs result from PyTorch's runtime-managed execution model: the Dynamo runtime scheduler performs specialization-cache lookup and dispatch, while Inductor independently compiles the host operations of each selected FX graph. Under this model, every invocation inevitably incurs the overhead of runtime-side specialization management, and even a cache hit must still pass through the runtime calls and glue code that connect Dynamo's selected specialization to Inductor's host wrapper.
In this paper, we investigate whether this recurring path can be removed by compiling guard-based specialization selection and preparation together with host execution.
This requires a compiled program that manages multiple guarded specializations while retaining the ability to return to Python when no specialization matches.

In this paper, we propose \textsc{Trident}, a compiler backend that reduces runtime overhead for programs containing user-written Triton kernels. 
Built on Torch-MLIR, \textsc{Trident} generates a \textit{Specialization Cache Module} (SCM), an executable module in which specialization management and host execution are compiled together for multiple guarded specializations (Figure~\ref{fig:intro-guard}(c)). In contrast to PyTorch's runtime-managed execution model, \textsc{Trident} routes every invocation directly into the SCM. The SCM evaluates the guards of its specializations, executes the corresponding host logic in the same module when a guard succeeds, and returns to Python only when no specialization matches.
This organization removes the runtime calls and glue code that connect specialization management to host execution on the cache-hit path.
To preserve the performance of existing PyTorch operators, ATen operations remain calls to their registered runtime implementations. 

Specifically, this paper makes the following contributions:
\begin{itemize}
    \item We introduce the \textit{Specialization Cache Module} (SCM), which compiles guard-based specialization selection, preparation of arguments and execution environment, and host execution for multiple specializations into a single executable module. The SCM eliminates the intervening runtime calls and glue code while retaining Python-side compilation when no specialization matches.

    \item We design and implement \textsc{Trident}, an end-to-end compiler backend built on Torch-MLIR. \textsc{Trident} analyzes Dynamo-captured FX graphs and guards, lowers the guards and host-side orchestration to C++ code, and exposes the resulting SCM through a TVM FFI entry. During lowering, \textsc{Trident} retains supported ATen operators as runtime calls to reuse their optimized implementations.

    \item We evaluate \textsc{Trident} on Triton operators and two LLMs, DeepSeek-V2-Lite and Qwen3-8B, across standard datasets. \textsc{Trident} achieves up to 1.73$\times$ end-to-end speedup over the eager wrapper on kernels and outperforms default \texttt{torch.compile} on most operators. At the model level, it achieves up to 1.47$\times$ and 1.68$\times$ end-to-end latency speedup over the eager wrapper and \texttt{torch.compile}, respectively, while also improving TPOT and ITL.
\end{itemize}


\section{Background and Motivation}
\label{sec:background}

\subsection{Triton Kernels}
\label{sec:bg-triton}

Triton is a Python-embedded DSL for writing GPU kernels~\cite{tillet2019triton}. Users write device computation in \texttt{@triton.jit} functions and call them from user host-side code. As illustrated in Figure~\ref{fig:intro-guard}(a), this user host-side code performs a series of operations such as buffer allocation, launch-argument preparation, kernel-variant selection, and kernel launch. Different inputs can lead to different host-side behavior. For example, the code may compute launch parameters from the input shapes, such as the grid size \(M\) and the compile-time constant \texttt{BLOCK\_SIZE}, and may select among kernel variants before issuing the kernel launch.

To avoid repeated compilation overhead, the Triton runtime caches compiled GPU binaries and reuses them across kernel calls~\cite{tillet2019triton}. For each call, it computes a cache key from properties of the invocation, such as argument types, \texttt{tl.constexpr} values, and launch options, and looks up this key in a per-device kernel cache. On a cache miss, Triton compiles the kernel to a GPU binary (CUBIN) and stores it under that key. On a hit, it reuses the cached CUBIN and issues the kernel launch. When \texttt{@triton.autotune} is used, the runtime additionally benchmarks a set of candidate configurations, such as the number of warps and the number of pipeline stages. It retains the fastest configuration in an in-memory map keyed by user-defined tuning attributes, and can persist these results to Triton's on-disk cache so that subsequent processes avoid re-autotuning. Later stages of \texttt{torch.compile} can reuse this cached CUBIN rather than recompiling the Triton kernel from source.

\subsection{Torch Compile}
\label{sec:bg-torch-compile}

The \texttt{torch.compile} compiles user host-side code while preserving the eager programming interface~\cite{pytorch2}. Dynamo captures graphs from Python execution. Inductor is the default compiler backend and generates a host wrapper for each captured FX graph.

\subsubsection{Dynamo Graph Capture and Guards}
\label{sec:bg-dynamo}

As shown in Figure~\ref{fig:intro-guard}(a), Dynamo executes and traces one path through the user's host-side Python code under a concrete runtime input~\cite{pytorch2}. In the \texttt{rms\_norm} example, different input shapes take different host-side branches. Dynamo extracts the tensor operations and Triton kernel calls along that executed path into an FX graph~\cite{reed2022torch}. During this capture, Dynamo records guards that describe the assumptions of the traced path. These guards include tensor-match checks on properties such as dtype, device, and shape-related metadata, control-flow and shape assumptions such as \texttt{weight.numel() <= 4096}. The resulting FX graph and its guards form a \emph{specialization}, which Dynamo caches for subsequent invocations.

On a later call, the Dynamo runtime scheduler performs a lookup in the specialization cache. A cache miss causes Dynamo to capture and compile a new specialization. PyTorch provides options that let users reduce this guard-related host overhead. A guard filter installed at compile time can drop selected guard kinds so that fewer checks are stored with each specialization, for example retaining tensor-match guards while discarding others. After warmup, users can further skip most remaining guard evaluation on later calls. Both options are unsafe, and a mismatch can silently produce incorrect results. Neither filtering nor skipping removes specialization-cache lookup, where the Dynamo runtime scheduler still walks the specialization cache to select the code to execute.

\subsubsection{Inductor Host Wrapper}
\label{sec:bg-inductor}

Inductor compiles each captured FX graph into a host wrapper~\cite{pytorch2}. At runtime, the selected specialization code performs argument preparation and environment preparation before calling the Inductor host wrapper. Argument preparation flattens the user-visible function arguments into the inputs expected by the selected FX graph and converts tensor arguments for the compiled calling interface. Environment preparation retrieves the current CUDA stream from the active device context. Host wrapper execution then performs memory management, launch-argument preparation, and kernel launch on that stream. For user-written Triton kernels, Inductor obtains the corresponding GPU binaries from Triton's kernel cache rather than recompiling those kernels from source.

By default, Inductor emits the wrapper as Python code. To accelerate host-wrapper execution, Inductor can emit a C++ wrapper that avoids Python interpretation overhead. It further supports capturing eligible GPU work within the wrapper as a CUDA Graph after warmup and replaying that graph on subsequent calls, which can reduce repeated kernel-launch cost within one specialization. CUDA Graph capture in Inductor remains subject to constraints on control flow and memory layout, and capture or replay may fail when these constraints are not satisfied. However, these mechanisms optimize only the host operations derived from a single FX graph. Neither the C++ wrapper nor CUDA Graph replay eliminates Dynamo's specialization-cache lookup or the argument and environment preparation that precedes wrapper entry.

\subsection{Motivation: Residual Host Overhead}
\label{sec:bg-motivation}

\begin{figure}[t]
  \centering
  \includegraphics[width=\columnwidth]{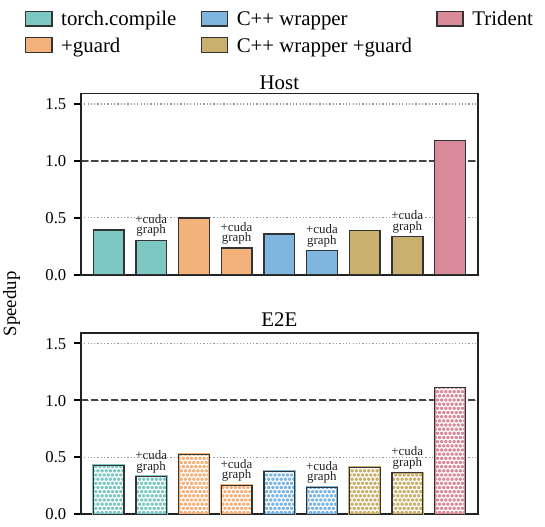}
  \caption{Warm-path host and end-to-end performance of \texttt{torch.compile} configurations on \texttt{rms\_norm}, normalized to the original eager wrapper.}
  \label{fig:motivation-rms-norm}
\end{figure}

Figure~\ref{fig:motivation-rms-norm} shows the warm-path host and end-to-end performance of representative \texttt{torch.compile} configurations on an \texttt{rms\_norm} example, normalized to the original Triton baseline with eager wrapper.
We run the experiment on an NVIDIA H800 GPU with input x of shape (1024, 128) and weight of shape (128,), warm up for 20 iterations, measure 10 repeats over 5 rounds for each configuration, and report the median.

Across these settings, every \texttt{torch.compile} configuration is slower than the eager baseline. The default configuration reaches only 0.39$\times$ of the eager baseline on the host and 0.43$\times$ end to end. Skipping guards improves this to 0.50$\times$ and 0.53$\times$ of the eager baseline, respectively, but remains well below eager. The C++ wrapper is even slower than default \texttt{torch.compile}, reaching only 0.36$\times$ and 0.38$\times$ of the eager baseline. Enabling CUDA Graphs is even slower, because replay-time parameter copies into static placeholders can outweigh the launch savings on short kernels~\cite{ghosh2026grace}. Note that we measure a single input shape after sufficient warmup, so this gap reflects residual host-side overhead on the warm path rather than cold compilation or shape switching. 

\begin{figure*}[t]
  \centering
  \includegraphics[width=\textwidth]{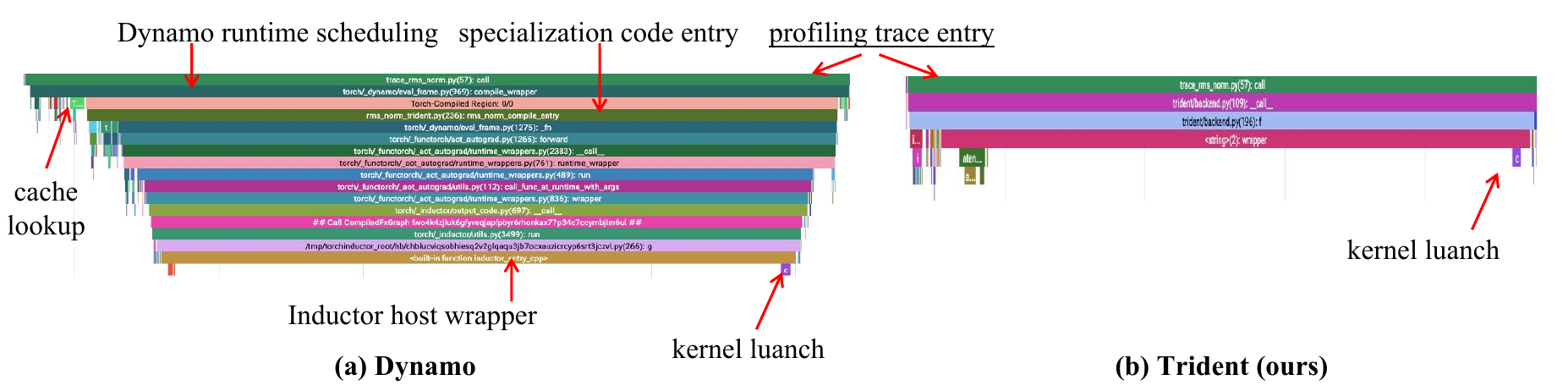}
  \caption{Warm-path Perfetto traces of \texttt{rms\_norm}: (a)~Dynamo with the Inductor C++ wrapper; (b)~\textsc{Trident}.}
  \label{fig:motivation-trace}
\end{figure*}

To further support this observation, we export a Perfetto~\cite{perfetto} trace of the warm-path C++ wrapper configuration, as illustrated in Figure~\ref{fig:motivation-trace}(a). Due to tracing overhead, these profiles are not accurate for absolute performance, but they show that a call reaches C++ wrapper execution only after many intervening runtime frames, which causes the residual host-side overhead and performance degradation.

To this end, we propose \textsc{Trident}, a compiler backend that generates a Specialization Cache Module (SCM) to eliminate the intervening runtime calls and glue code on the warm path. Figure~\ref{fig:motivation-rms-norm} shows that \textsc{Trident} is the only configuration that improves over the eager baseline, with 1.18$\times$ host and 1.11$\times$ end-to-end speedup, and that it is 2.99$\times$ and 2.60$\times$ faster than default \texttt{torch.compile} on host and end-to-end latency, respectively. Figure~\ref{fig:motivation-trace}(b) further shows that \textsc{Trident} reaches kernel launch with far fewer intervening runtime frames than the Dynamo path.

\section{Methodology}
\label{sec:methodology}

\subsection{Overview}

\textsc{Trident} requires no changes to the body of a supported Python function or its user-written Triton kernels: users simply apply \texttt{@trident.jit} to the Python entry function.
Existing PyTorch operations and Triton kernel launches remain unchanged, and the kernels retain their \texttt{@triton.jit} decorators.
From this function, TorchDynamo captures an ATen-level FX graph together with the guards that define when the captured computation is valid.
\textsc{Trident} turns this guarded input into a compiled multi-version program that handles the recurring cache-hit path.
The output is a callable that preserves the source-level signature while routing each invocation to a compatible compiled specialization.
Unlike a conventional per-graph host wrapper, this callable moves specialization selection, graph-input preparation, and host execution into the same compiled boundary.

The central artifact is the \emph{Specialization Cache Module} (SCM), illustrated in Figure~\ref{fig:intro-guard}(c).
An SCM contains an ordered set of versions, where each version pairs a guard predicate with the computation captured under that predicate.
It also contains a dispatcher that evaluates these predicates and transfers control to the first matching computation.
The Python boundary is deliberately narrow: it prepares source arguments for the compiled calling convention and initiates compilation only when the SCM reports that no version matches.
Consequently, a cache hit crosses from Python into the SCM once and remains in compiled host code through version selection and execution.
Calls from the SCM to registered ATen implementations are internal runtime calls and do not return control to the Python scheduler.

The design is governed by three requirements:
\begin{enumerate}
  \item \textbf{Safe reuse.}
  Every assumption that determines the captured Python path or a selected Triton CUBIN must hold before the operation it protects executes.
  \item \textbf{Stable interface.}
  All versions must expose the source-level calling convention despite capture-specific input flattening and internal signatures.
  \item \textbf{Runtime reuse.}
  Compilation should handle ATen and Triton calls with little or no change to supported Python programs while reusing their existing implementations.
\end{enumerate}
The SCM meets these requirements by pairing each version's guards with its computation, wrapping all versions behind one source-level interface, and reusing PyTorch's ATen implementations and Triton's selected CUBINs.

Compilation proceeds through a series of representation transitions rather than by rewriting the user's kernels.
TorchDynamo first captures an ATen-level FX graph and the assumptions of the executed Python path.
For each capture, \textsc{Trident} packages the FX graph, its validity checks, and any selected Triton CUBINs into a guarded high-level MLIR submodule, which it adds to the stored versions.
It then rebuilds the executable SCM by merging all stored submodules, lowering the combined module to the LLVM dialect, adding an ordered dispatcher, and JIT-compiling the complete module in an in-process execution engine.
Section~\ref{sec:guards} explains native guarded dispatch, Section~\ref{sec:fx-to-torch-mlir} describes how a specialization is built, Section~\ref{sec:host-codegen} explains how the SCM is lowered and installed, and Section~\ref{sec:execution-lifecycle} describes runtime execution and specialization growth.

\subsection{Native Guarded Dispatch}
\label{sec:guards}

In \texttt{torch.compile}, guard management and graph-module execution remain separate runtime stages.
As Figure~\ref{fig:intro-guard}(b) shows, the Dynamo runtime scheduler owns specialization-cache lookup and guard evaluation, whereas a backend-generated host wrapper owns execution of the captured FX GraphModule.
Even on a cache hit, glue code must connect these stages by preparing graph inputs and the execution environment before transferring control to the wrapper.
Because the stages span several language and runtime boundaries, one invocation can require multiple cross-boundary calls.
A C++ wrapper accelerates execution inside the selected graph but does not remove the preceding runtime work or boundary crossings.
For short GPU kernels, the fixed overhead of this glue code can lie on the critical path.

As Figure~\ref{fig:intro-guard}(c) shows, \textsc{Trident} unifies these stages in the SCM.
It places each version's guards and captured graph computation in one module.
During lowering, the private graph computation is inlined into its guarded wrapper, so guard control flow and graph host operations become one native program.
A guarded specialization is therefore the unit of both reuse and correctness: once an invocation enters the SCM through one TVM FFI entry, native control flow evaluates guards, advances across rejected versions, and enters the first matching graph computation without returning to the Python scheduler between these steps.
For an input $x$, let $G_i(x)$ denote the assumptions of version $i$ and let $P_i(x)$ denote its captured computation.
Algorithm~\ref{alg:scm-dispatch} formalizes ordered dispatch. The first version whose assumptions hold executes; if every $G_i$ fails, dispatch returns a specialization miss (lines 2-3, 8).
An execution error from the selected $P_i$ terminates dispatch immediately and propagates through TVM FFI's error channel rather than being treated as a miss (lines 5-6).
At the Python callable boundary, TVM FFI raises the corresponding exception to the application, so \textsc{Trident} neither tries another version nor triggers recompilation.
This organization removes repeated boundary crossings between specialization management and graph execution on the cache-hit path.
Necessary internal runtime calls, such as ATen dispatch through Trident FFI, remain within the matched SCM and do not re-enter Python-side specialization management.

\begin{algorithm}[t]
  \caption{Ordered dispatch in an SCM.}
  \label{alg:scm-dispatch}
  \SetKw{Continue}{continue}
  \SetKw{Raise}{raise}
  \KwIn{Input $x$; ordered versions $\{(G_i, P_i)\}_{i=1}^{n}$}
  \KwOut{Program result or \textsc{Miss}}
  \For{$i \leftarrow 1$ \KwTo $n$}{
    \If{$G_i(x)$ fails}{
      \Continue\;
    }
    $o \leftarrow P_i(x)$\;
    \If{$o = \textsc{Error}(e)$}{
      \Raise{$e$}\;
    }
    \Return{$o$}\;
  }
  \Return{\textsc{Miss}}\;
\end{algorithm}

\begin{figure*}[t]
  \centering
  \includegraphics[width=\textwidth]{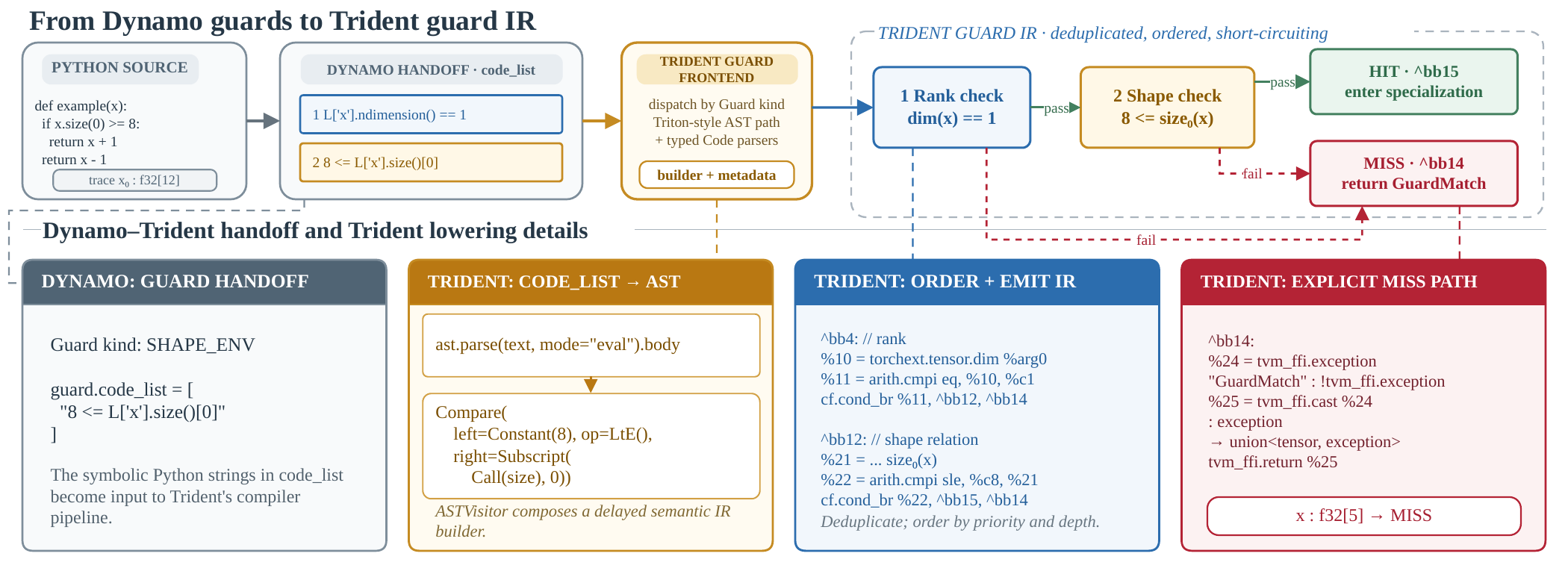}
  \caption{Lowering the assumptions of a captured Python path into guard IR.
  TorchDynamo supplies the captured assumptions, which \textsc{Trident} converts into dependency-ordered semantic checks.
  The resulting short-circuiting control-flow graph enters the captured computation only after all checks succeed and otherwise returns an explicit specialization miss.}
  \Description{A guard-lowering overview for a Python function whose branch depends on the length of a one-dimensional tensor.
  The captured assumptions are translated into a rank check followed by a shape-bound check.
  Passing both checks enters the compiled specialization, whereas either failure returns a specialization miss.
  \shaohua{add @trident.jit in PYTHON SOURCE}}
  \label{fig:guard-ir}
\end{figure*}

To implement $G_i(x)$, \textsc{Trident} translates the supported TorchDynamo guards associated with each captured version into checks over source-level runtime inputs.
It combines predicate-derived checks with checks for other captured assumptions, removes redundancies, and orders the remaining checks according to their data dependencies.
The resulting short-circuiting control flow enters $P_i(x)$ only after every required check succeeds; any failed check reports a specialization miss.
The lowering preserves symbolic relationships that can remain valid across multiple concrete inputs.
Guard evaluation and graph-input preparation resolve values through the same logical parameter mapping, ensuring that each predicate protects the corresponding captured computation.
Keeping guards explicit in compiler IR allows them to be optimized and compiled together with the computation they protect.
Figure~\ref{fig:guard-ir} summarizes this translation from captured assumptions to compiled guard control flow.

\subsection{Building a Specialization}
\label{sec:fx-to-torch-mlir}

Native guarded dispatch requires every captured computation to remain paired with the assumptions that make it safe to reuse and to expose an interface shared by all versions.
\textsc{Trident} therefore packages each guarded specialization as a self-contained MLIR submodule, keeping its validity checks and captured work together while allowing it to be merged with the stored versions when rebuilding the SCM.

Construction begins with graph capture and normalization.
TorchDynamo captures the executed path as an ATen-level FX graph with its guards.
\textsc{Trident} preserves Dynamo's capture assumptions and symbolic values while normalizing the graph into the forms expected by the importer.
Torch-MLIR~\cite{torchmlir} then imports the normalized graph as a private function in the new submodule.
Giving every imported computation and wrapper a version-specific symbol keeps captures independent when their submodules are later merged.

The exported graph and the source function intentionally use different input representations.
The Python boundary sees structured source arguments, whereas the imported graph consumes a flat sequence of graph leaves.
Flattening is convenient for graph transformations, but exposing it at the external boundary would require Python to choose a specialization before it knew which flat signature to construct.
\textsc{Trident} instead arranges each version wrapper's parameters in the original function's parameter order and reconstructs the graph inputs only after that version's guards succeed.

This reconstruction is driven by the exported input tree and graph signature.
For each source parameter, \textsc{Trident} records a typed recipe describing its source-level structure.
At wrapper entry, the recipe binds the actual argument to a logical input tree.
Guard expressions address values through paths in this tree, while the success block recursively flattens the same tree in exporter order and selects the leaves consumed by the imported computation.
Structured values are unpacked only after the relevant guards establish that doing so is valid.
This shared reconstruction mechanism aligns a guard, such as a check on the second tensor in a tuple, with the corresponding graph operand.

Source-visible constants that affect specialization but are absent from the flattened graph remain wrapper parameters and can participate in guards.
Values embedded during export remain part of the captured computation instead of being recreated at runtime.
On return, result normalization restores the Python-facing representation.
Thus, the external signature remains stable even when versions differ internally in their flattened operands or result structure.

Each Triton call additionally requires a launch description and the CUBIN selected for the current specialization.
During capture, executing the exported graph causes Triton to compile or retrieve the kernel variant determined by the current inputs and compile-time launch configuration.
\textsc{Trident} obtains the selected CUBIN from Triton's device-local cache and embeds it in the submodule under a version-specific symbol.
The high-level MLIR submodule pairs this binary with a TorchExt launch operation that refers to its symbol and records the launch description and specialization requirements.
This representation reuses Triton's compiled device code without recompiling the Triton program from source.

During whole-SCM lowering, \textsc{Trident} converts the launch description into GPU IR while retaining the embedded CUBIN.
It materializes the kernel's specialization checks before the launch and converts the runtime arguments to the kernel's native calling convention.
The resulting GPU launch operation carries the dynamic launch dimensions and refers to the same binary.
Lowering and translation to LLVM then produce compiled host code that loads the embedded GPU module, resolves the kernel entry, and launches it on the current stream.

Crucially, this construction removes the Triton Python runtime from the recurring cache-hit path.
Later hits validate and invoke the embedded CUBIN from compiled host code, bypassing Triton's Python specialization and launch path.
The cost of Triton runtime specialization is therefore incurred once when each specialization is built, rather than on every invocation.

\begin{figure*}[t]
  \centering
  \includegraphics[width=\textwidth]{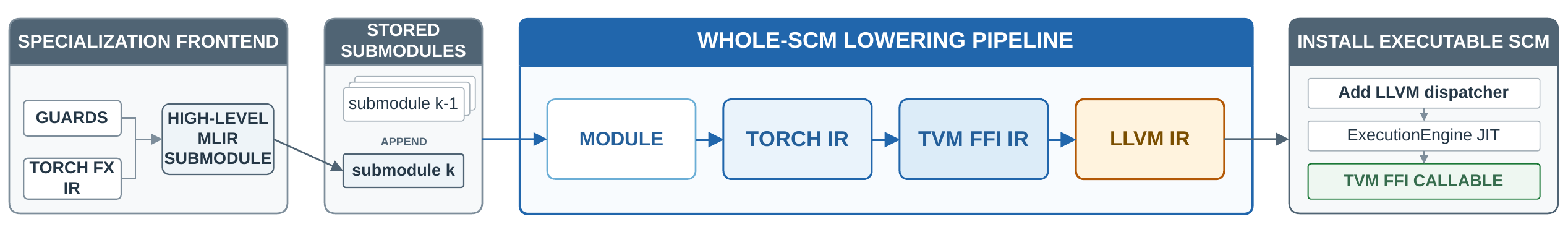}
  \caption{Whole-SCM compilation centered on the main host-lowering path.
  Each capture contributes a high-level MLIR submodule to the stored versions.
  \textsc{Trident} merges these submodules and lowers the combined module along the main host path from Torch IR through TVM FFI IR to the LLVM dialect.
  It then adds the LLVM dispatcher and JIT-compiles the executable SCM behind one TVM FFI callable.}
  \Description{A compilation diagram dominated by the whole-SCM lowering pipeline.
  A compact frontend converts guards and Torch FX IR into a high-level MLIR submodule and appends it to the stored versions.
  The enlarged central pipeline merges all versions and highlights the path from Torch IR through TVM FFI IR to the LLVM dialect over the combined module.
  A compact final stage installs the lowered module behind one TVM FFI callable.}
  \label{fig:trident-lowering-pipeline}
\end{figure*}

\subsection{SCM Lowering and Runtime Integration}
\label{sec:host-codegen}

Building an individual specialization yields a reusable high-level submodule, but an executable SCM must combine and lower all stored versions into one program with an ordered dispatcher.
\textsc{Trident} therefore constructs the SCM at two levels: the frontend builds one high-level submodule for each specialization, whereas an SCM rebuild lowers all stored versions together.
Figure~\ref{fig:trident-lowering-pipeline} distinguishes these two levels explicitly.
Before a rebuild, every stored specialization remains a high-level MLIR submodule that retains its source-level semantics.
\textsc{Trident} combines these submodules and applies one whole-SCM lowering pipeline, which inlines each graph and converts the combined host program through TVM FFI IR to the LLVM dialect.
The same pipeline lowers Triton launch descriptions and preserves every specialization as a distinct lowered entry.

ATen operators use a compiled calling boundary that reuses PyTorch's registered implementations.
We provide \emph{Trident FFI}, a schema-driven binding layer that covers the registered ATen operator surface of the installed PyTorch build and generates packed wrappers for it.
We implement these wrappers with a generic C++ template instead of handwritten per-operator bindings.
At build time, a Python generator enumerates the ATen schemas registered with PyTorch's dispatcher, maps their argument and result types to template parameters, and emits one corresponding wrapper instantiation for each operator.
Shared template specializations translate between TVM FFI values and PyTorch values, while the generic wrapper invokes the registered implementation through PyTorch's boxed dispatcher and returns its result to the SCM.
When the generated library is loaded, Trident FFI registers these wrappers under stable \texttt{trident.aten.*} names in TVM FFI's global function table.
For a general ATen operator, lowering performs a name-based lookup in this table and emits a packed call to the resolved wrapper.
The compiler can therefore lower an ATen operation without embedding PyTorch's dispatcher data structures into its own IR.

\textsc{Trident} divides ATen-related computation between two paths:
\begin{itemize}
  \item \emph{Native lowering.} Selected inexpensive host operations, such as guard arithmetic and tensor-metadata queries, become native instructions when their semantics and ABI representation are known.
  These operations commonly occur in guards and launch-grid calculations, where a runtime operator lookup would be disproportionately expensive.
  \item \emph{Runtime dispatch.} General tensor operators call their generated Trident FFI entries and reuse the implementations registered with PyTorch.
  An FFI and dispatcher boundary remains for each such operator, but Python no longer schedules these operations or connects them to the selected specialization.
\end{itemize}
This split targets orchestration overhead while retaining the coverage and device-specific behavior of the existing ATen runtime.

Triton launches follow a separate lowering path because their computation is already present as a CUBIN.
For every launch, the host code first evaluates the CUBIN-specific preconditions described in Section~\ref{sec:fx-to-torch-mlir}.
It then prepares the runtime arguments and launch configuration before submitting the embedded CUBIN to the current CUDA stream.
Symbolic shape and grid arithmetic remains in compiler IR throughout this path, allowing constant fragments to be folded while input-dependent fragments remain native runtime calculations.
The selected Triton kernel therefore keeps its original CUBIN while its repeated launch preparation becomes part of the compiled SCM.

The TVM FFI representation provides the common ABI across SCM and runtime boundaries.
At compile time, lowering maps Torch-level tensor and scalar types, together with the operations that consume them, to typed semantic TVM FFI IR; a subsequent conversion lowers this representation to the packed TVM FFI ABI.
At execution time, TVM FFI automatically converts and packs Python arguments at the callable boundary, representing PyTorch tensors through DLPack.
Runtime function lookup and invocation are checked explicitly, so a missing registration becomes a real execution error.
Normal program values and the specialization-miss object share a general result slot, but their runtime type tags remain distinct.
This distinction lets the LLVM dispatcher identify a miss without reserving an error code that could be confused with an operator failure.

Finally, \textsc{Trident} lowers the combined module's remaining control flow and ABI operations to the LLVM dialect.
After this lowering pipeline completes, it adds the ordered LLVM dispatcher under one stable symbol derived from the original function name.
The execution engine optimizes and JIT-compiles the complete LLVM module, then exposes its entry symbol as a TVM FFI callable.
The callable retains the engine so that all executable state remains live for the lifetime of the callable.

\subsection{Runtime Execution and Specialization Growth}
\label{sec:execution-lifecycle}

The installed SCM separates the frequent compiled hit path from the infrequent Python compilation path while distinguishing specialization misses from execution errors.
On each call, the outer wrapper binds the source arguments and normalizes them into their FFI representations.
It then makes one entry call from Python into the SCM dispatcher.
This boundary performs representation normalization but does not inspect Dynamo's specialization cache or choose a captured graph.

On a cache hit, the dispatcher tries versions in creation order.
For each version, its wrapper first evaluates the short-circuiting Dynamo-derived guards.
A failed guard returns the distinguished miss value, causing the dispatcher to advance to the next version without unwinding through Python.
After all guards succeed, the wrapper reconstructs the flattened graph operands and enters the inlined host computation; launch-specific preconditions are checked immediately before the corresponding Triton CUBIN is submitted.
The result travels through the same FFI boundary and is normalized back into its source-level Python representation.
The entire cache-hit path therefore remains on the compiled side of the single Python-to-SCM entry.

On a cache miss, every installed wrapper returns the distinguished miss value.
The dispatcher forwards the final miss to the outer wrapper, which invokes Python only at this point to capture a new specialization and rebuild the SCM.
An empty SCM follows the same path on its first invocation.
The invocation that triggered compilation returns the result produced during capture, while later compatible invocations use the newly installed compiled version.
This policy incurs the cost of whole-module reconstruction when the specialization set grows in exchange for a simpler steady-state path with lower overhead.
\textsc{Trident} then rebuilds and JIT-compiles a replacement SCM from the complete stored version set under the same public entry point.

Real execution errors form a third path and never trigger specialization growth.
If an internal runtime call fails, the generated FFI code propagates its error status to the dispatcher, which terminates rather than attempting another version.
This behavior preserves the semantic difference between ``the assumptions of this version do not hold'' and ``the selected program failed to execute.''
Together, the three paths give the SCM a precise responsibility boundary: compiled code handles expected input variation among installed versions, whereas Python handles only previously unseen variation that requires a new capture.

\section{Evaluation}
\label{sec:evaluation}

\subsection{Experiment Setup}

\textbf{Platform --}
We conduct all experiments on a server with two Intel Xeon Platinum 8468H processors (96 physical cores and 192 hardware threads) and NVIDIA H800 GPUs with 80\,GB of HBM. Each benchmark uses a single GPU. The system runs NVIDIA driver 580.167.08 and PyTorch 2.11.0.

\textbf{Kernel benchmarks --}
We use operators from FlagGems~\cite{flaggems2024}, a widely used open-source Triton operator library in the PyTorch ecosystem. We select 21 representative operators that span the main computational idioms in LLM and deep-learning kernels: pointwise arithmetic and activations (e.g., \textit{add}, \textit{silu}, \textit{pow}), comparisons and masking (e.g., \textit{lt}, \textit{masked\_fill}), normalization (\textit{rms\_norm}), embedding lookup (\textit{embedding}), scans (\textit{cumsum}), sorting (\textit{sort}), dense matrix multiplication (\textit{mm}, \textit{addmm}), concatenation (\textit{cat}), and convolution (\textit{conv2d}, \textit{conv\_transpose2d}). For each operator, we select five representative shapes derived from common LLM tensor dimensions. We time 30 consecutive invocations per shape. We report the first as cold-start latency and the median of the remaining invocations as warm latency. We report both end-to-end latency (including host and device execution until GPU completion) and host latency (only until the Python call returns).

\textbf{Model benchmarks --}
We evaluate DeepSeek-V2-Lite and Qwen3-8B~\cite{yang2025qwen3} on MMLU (general knowledge)~\cite{hendrycks2021mmlu}, GSM8K (mathematical reasoning)~\cite{cobbe2021gsm8k}, and HumanEval (code generation)~\cite{chen2021humaneval}. All experiments use batch size one and generate at most 128 new tokens per request. We compare the eager Python wrapper, \texttt{torch.compile}, \texttt{torch.compile} with the C++ wrapper, and Trident while replacing the same operator whitelist---\textit{bmm}, \textit{linear}, and \textit{embedding}---in each model. Because wrapper behavior on real model inputs is complex, we do not enable additional aggressive acceleration options that would compromise correctness, and CUDA Graphs fail without model-specific adaptation. For every mode and dataset, we first execute ten warmup requests and then measure 64 requests for DeepSeek-V2-Lite and 32 requests for Qwen3-8B.

\subsection{Kernel Performance}

\begin{figure*}[t]
  \centering
  \includegraphics[width=\textwidth]{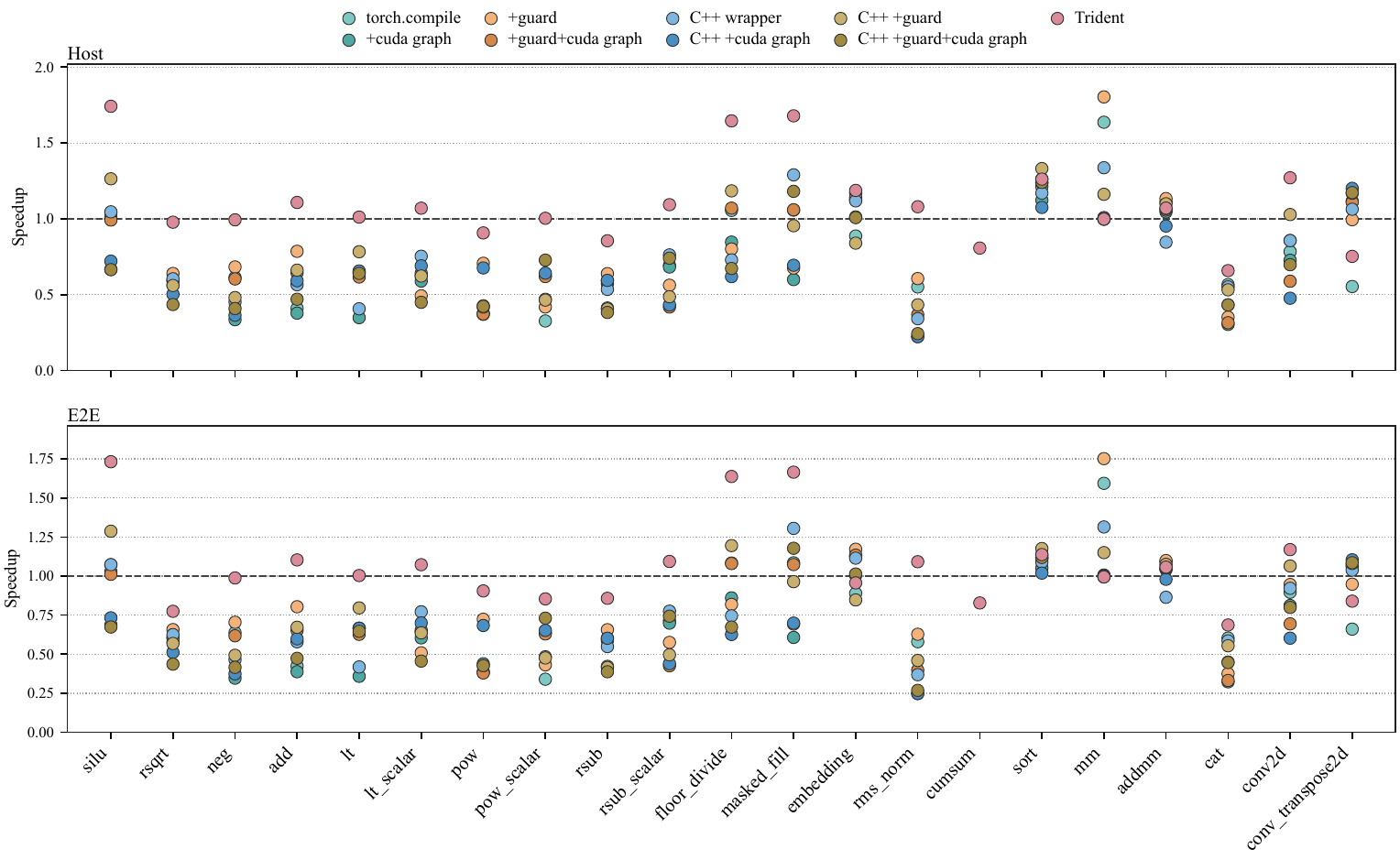}
  \caption{Warm host and end-to-end performance relative to the eager wrapper, averaged over five representative shapes per operator. Higher is better; values below $1\times$ are slower than eager. Missing points correspond to configurations that failed to run.}
  \label{fig:kernel-h800-speedup}
\end{figure*}

Figure~\ref{fig:kernel-h800-speedup} reports warm-path performance relative to the eager wrapper. For each operator and execution mode, we discard the first timed invocation, remove outliers from the remaining samples using the 1.5-IQR rule, and compute the mean latency; the figure reports the arithmetic mean of the five per-shape ratios. Relative to the eager wrapper, \textsc{Trident} is faster on most operators (13 of 21 in host latency and 11 of 21 end to end), with mean speedups of 1.10$\times$ and 1.07$\times$. More importantly, on operators with a complete \texttt{torch.compile} baseline, \textsc{Trident} is faster than default \texttt{torch.compile} on 18 of 20 operators, by up to 3.07$\times$ in host latency and 2.61$\times$ end to end, and it likewise beats the other evaluated \texttt{torch.compile} configurations on most of the suite.

The largest gains occur when a short device kernel is surrounded by substantial host-side dispatch and launch preparation. For example, \textsc{Trident} reaches up to 1.74$\times$ host and 1.73$\times$ end-to-end speedup on \textit{silu}, and 1.67$\times$ end-to-end on \textit{masked\_fill}: by placing specialization dispatch, argument preparation, and host execution in one SCM, it removes repeated wrapper orchestration that still remains on the eager and \texttt{torch.compile} warm paths. Conversely, operators such as \textit{cat} remain below 1$\times$ when memory movement or device work dominates, or when too much wrapper logic stays outside the SCM.

This pattern generalizes the \textit{rms\_norm} observation in Section~\ref{sec:bg-motivation}. Compiling an FX graph removes repeated interpretation of the original wrapper, but a warm call still enters the Dynamo scheduler for specialization-cache lookup and guard evaluation, performs argument and environment preparation, and crosses into the Inductor wrapper. A C++ wrapper or CUDA-Graph replay optimizes only work inside the selected graph and does not remove that preceding runtime path, so \texttt{torch.compile} is often slower than the already lean eager wrapper. By compiling guarded dispatch and graph execution within the same SCM, \textsc{Trident} removes this residual overhead and is therefore faster than \texttt{torch.compile} on most operators.

Cold start is slower. The first timed invocation includes Triton compilation and, for \textsc{Trident}, SCM construction, so its cold host latency is typically higher than \texttt{torch.compile}---about 2$\times$ across many short kernels, e.g., 0.84\,s versus 0.42\,s on \textit{neg}. This cost is paid during warmup. Once the specialization is cached, repeated warm invocations recover the overhead through the per-call savings above. Although the speedup of a single operator invocation is often modest, model inference repeatedly invokes the same kernels across layers and autoregressive decoding steps, allowing these per-invocation savings to accumulate at model scale.

\subsection{Model Performance}

\begin{figure*}[t]
  \centering
  \includegraphics[width=\textwidth]{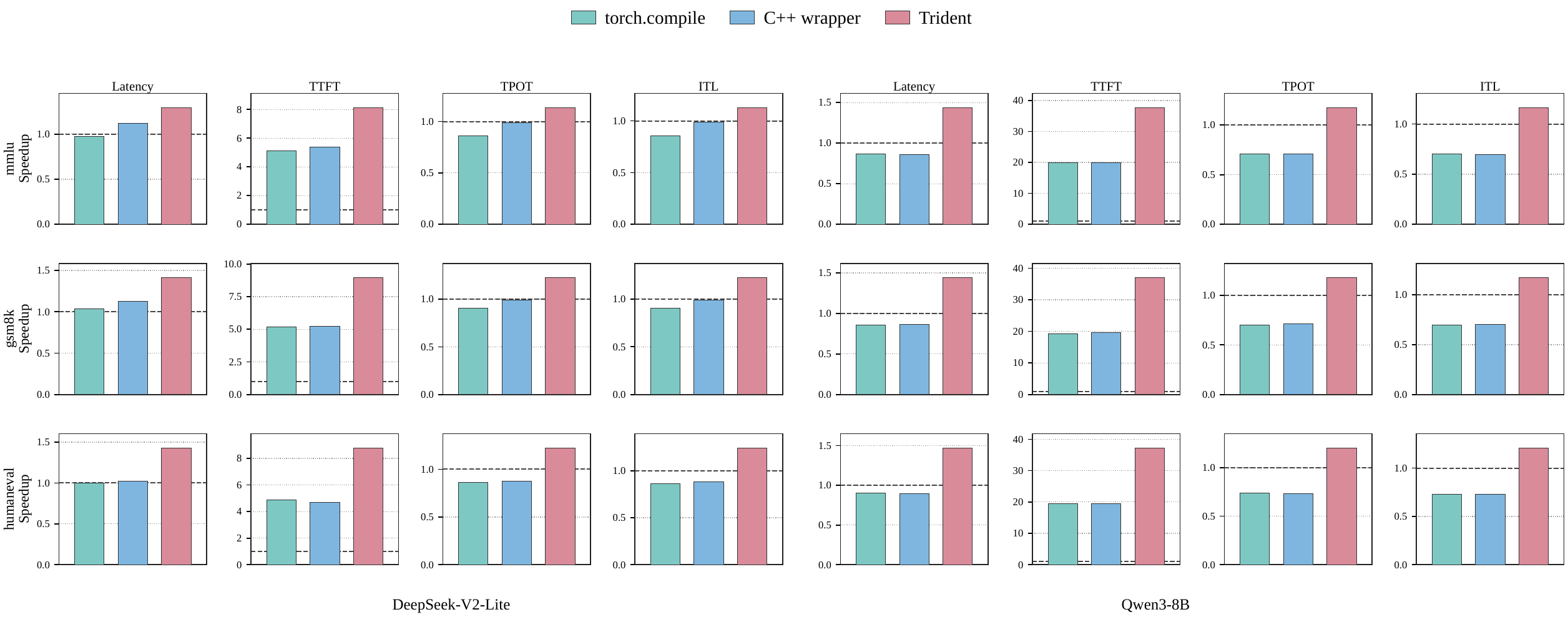}
  \caption{Model-level median performance relative to the eager Python wrapper on an NVIDIA H800. Results cover DeepSeek-V2-Lite and Qwen3-8B on MMLU, GSM8K, and HumanEval. Higher is better; values below $1\times$ are slower than eager.}
  \label{fig:model-h800-speedup}
\end{figure*}

Figure~\ref{fig:model-h800-speedup} reports model-level performance relative to the eager Python wrapper. Whether compiling an operator is beneficial is difficult to determine statically. We therefore use ten warmup requests to identify and exclude operator candidates that exhibit negative optimization, and apply this filtering policy uniformly to all compilation-based configurations, including \texttt{torch.compile}, its C++ wrapper, and \textsc{Trident}. Across all six model--dataset combinations, \textsc{Trident} consistently improves every reported median metric: it achieves 1.30--1.47$\times$ end-to-end latency speedup, 1.14--1.23$\times$ TPOT speedup, and 1.13--1.24$\times$ ITL speedup. In contrast, \texttt{torch.compile} reaches only 0.86--1.03$\times$ of the eager baseline in end-to-end latency and 0.70--0.90$\times$ in TPOT, while the C++ wrapper reaches 0.86--1.13$\times$ and 0.71--0.99$\times$ of the eager baseline, respectively. Thus, conventional compilation frequently regresses relative to the already optimized eager Python wrapper, whereas \textsc{Trident} turns host compilation into consistent end-to-end gains.

The improvement is consistent across both model architectures and all three workload domains. For \texttt{torch.compile}, \textsc{Trident} achieves 1.33--1.68$\times$ median end-to-end latency speedup; relative to the C++ wrapper, it achieves 1.16--1.67$\times$. The corresponding TPOT gains are 1.32--1.68$\times$ and 1.15--1.67$\times$ over \texttt{torch.compile} and C++ wrapper, respectively. These results show that compiling only the selected graph wrapper is insufficient to reliably outperform the eager Python wrapper at model scale. By placing specialization dispatch and host execution in the same SCM, \textsc{Trident} preserves the optimized device kernels while reducing the residual runtime overhead around them. On Qwen3-8B, the eager Python wrapper is particularly slow on the first token: prefill issues a long sequence of large \textit{linear}/\textit{bmm} calls through the Python host path, so wrapper overhead accumulates before the first output token is produced. We therefore emphasize end-to-end latency together with the decode-path metrics TPOT and ITL, which better reflect steady-state serving behavior.

\section{Related Work}
\label{sec:relatedwork}

\paragraph{Reducing host-side GPU orchestration.}
Prior systems reduce CPU--GPU orchestration costs by changing how device work is submitted or scheduled.
CUDA Graphs capture a fixed workflow and replay it with one graph launch; PyTorch applies this mechanism in its reduce-overhead mode, while GraCE uses compiler transformations, parameter-copy elimination, and cost--benefit analysis to increase the workloads for which graph replay is profitable~\cite{pytorch2,ghosh2026grace}.
Other systems move scheduling further onto the device.
Rammer constructs a static spatio-temporal schedule across and within operators~\cite{ma2020rammer}.
Mega-kernel systems push this direction further by executing multiple operators within one device-resident kernel, thereby avoiding repeated host submissions.
MonoNN compiles a static single-GPU neural network into a monolithic kernel, while FlashMoE fuses the computation and communication of a distributed MoE layer into one persistent kernel~\cite{zhuang2024mononn,aimuyo2025flashmoe}.
MPK automates model-level mega-kernelization by lowering multi-GPU inference to an SM-level task graph executed by an in-kernel runtime~\cite{cheng2026mpk}; Event Tensor further encodes fine-grained task dependencies while supporting dynamic shapes and data-dependent execution in mega-kernels~\cite{jin2026eventtensor}.
These approaches reduce launch gaps by changing the granularity and location of device scheduling.
In contrast, \textsc{Trident} preserves existing device-kernel boundaries and compiles the framework-side guards, specialization selection, and host execution that precede and connect their launches.

\paragraph{Host code generation.}
The closest systems compile host logic around device kernels.
TorchInductor generates a host wrapper for each captured FX graph to perform tasks such as tensor-size computation, memory management, and calls to generated or external kernels; its C++ wrapper reduces the interpretation overhead of the default Python wrapper~\cite{pytorch2}.
However, Dynamo still performs specialization-cache lookup and guard evaluation before entering that wrapper.
Dynamic-shape compilers generate a broader runtime path: DISC compiles shape inference, buffer management, kernel-launch control, and device computation together, while BladeDISC combines multiversion code generation with runtime selection of shape-appropriate kernels~\cite{zhu2021disc,zheng2023bladedisc}.
TVM FFI provides a compact calling convention and zero-copy tensor interoperability for connecting compiled kernels to framework runtimes~\cite{tvmffi}.
TileLang provides a controllable language and compiler for fused neural kernels~\cite{wang2026tilelang}.
In the DeepSeek-V4 deployment stack, Host Codegen complements these kernels by co-generating a lightweight host launcher that validates tensor contracts and marshals arguments through TVM FFI~\cite{deepseek2026v4}.
This launcher removes Python from the per-kernel validation path.
These systems generate host logic within one graph execution or around an individual kernel, whereas \textsc{Trident} compiles guard-based selection and host execution across multiple Dynamo specializations into one SCM.

\paragraph{Compiling dynamic eager programs.}
Work on imperative and dynamic neural programs has primarily focused on determining which computation can be represented and optimized as a graph.
JANUS speculatively converts Python control flow, dynamic types, and side effects into a symbolic graph~\cite{jeong2019janus}; Nimble represents dynamic control flow, data structures, and tensor shapes with a dynamic type system and a lightweight virtual machine~\cite{shen2021nimble}; and MAGPY monitors execution state and reference relationships to capture more complete operator graphs from eager programs~\cite{zhang2024magpy}.
Within PyTorch 2's \texttt{torch.compile} stack, TorchDynamo symbolically analyzes Python bytecode to extract FX graphs and associated guards, reusing a compiled specialization when its guards hold~\cite{pytorch2}.

\section{Conclusion}
\label{sec:conclusion}

This paper studies host-side overhead in PyTorch programs that invoke user-written Triton kernels. For short kernels, specialization management and host preparation outside compiled execution can dominate end-to-end latency. We propose \textsc{Trident}, a Torch-MLIR-based compiler backend that addresses this cost with the \emph{Specialization Cache Module} (SCM). The SCM compiles guarded specialization selection, preparation, and host execution into a single executable module, so a cache hit stays in compiled code and returns to Python only when a new specialization must be compiled. Our evaluation shows that \textsc{Trident} substantially reduces end-to-end latency on Triton operators and model workloads, indicating that compiling specialization management together with host execution is an effective way to realize the performance of customized Triton kernels.

\bibliographystyle{ACM-Reference-Format}
\bibliography{reference}

\end{document}